\documentclass[conference,a4paper]{IEEEtran}
\IEEEoverridecommandlockouts
\usepackage{cite}
\usepackage{amsmath,amssymb,amsfonts}
\usepackage{algorithmic}
\usepackage{graphicx}
\usepackage{textcomp}
\usepackage{xcolor}
\usepackage{bbm}
\usepackage{booktabs}
\usepackage{multirow}
\usepackage{url}
\def\BibTeX{{\rm B\kern-.05em{\sc i\kern-.025em b}\kern-.08em
    T\kern-.1667em\lower.7ex\hbox{E}\kern-.125emX}}
\begin{document}

\title{Sparsity-Adaptive Sharpness-Aware Minimization}
\author{\IEEEauthorblockN{Shiryu Ueno}
\IEEEauthorblockA{\textit{Faculty of Engineering} \\
\textit{Gifu University}\\
Gifu, Japan \\
ueno@cv.info.gifu-u.ac.jp}
\and
\IEEEauthorblockN{Yoshikazu Hayashi}
\IEEEauthorblockA{\textit{Faculty of Engineering} \\
\textit{Gifu University}\\
Gifu, Japan \\
hayashi.yoshikazu.a8@f.gifu-u.ac.jp}
\and
\IEEEauthorblockN{Kunihito Kato}
\IEEEauthorblockA{\textit{Faculty of Engineering} \\
\textit{Gifu University}\\
Gifu, Japan \\
kato.kunihito.k6@f.gifu-u.ac.jp}
}

\maketitle

\begin{abstract}
Deploying deep neural networks in real-world settings requires models that are both compact and robust to common corruptions. 
However, at deployment-relevant high sparsity, standard pruning pipelines often degrade corruption robustness, and existing sharpness-aware training/pruning approaches provide limited robustness gains.
We address this issue by introducing Sparsity-Adaptive Sharpness-Aware Minimization (SA-SAM), which derives a sparsity-dependent SAM/ASAM perturbation radius by keeping the mean absolute perturbation (an $\ell_1$-based proxy) approximately invariant as sparsity increases.
As a simple complementary option, we evaluate Magnitude-Weighted Hessian (MWH), derived from a second-order removal-path analysis, yielding an importance proportional to $\mathrm{Diag}(F)_i\,|w_i|$, where $\mathrm{Diag}(F)$ is the diagonal empirical Fisher used as a curvature proxy in our implementation.
Across CIFAR-10-C, CIFAR-100-C, and ImageNet-100-C, our approach achieved stronger corruption robustness than the considered pruning baselines at 80--90\% sparsity, while preserving clean accuracy.
We additionally quantify the robustness--throughput trade-off by reporting measured inference throughput under sparse execution at deployment-relevant sparsity levels.
\end{abstract}


\section{Introduction}
%
%
Pruning~\cite{cheng_survey_2024} is a major approach to model compression that removes parameters deemed less important by an importance score.
While most pruning methods primarily target preserving clean accuracy, they can compromise robustness to corruptions that frequently arise in deployment (e.g., the CIFAR-C/ImageNet-C setting~\cite{hendrycks_benchmarking_2018}).
Recent efforts therefore incorporate robustness into compression. 
For example, AdaSAP~\cite{bair_adaptive_2023} leverages Adaptive Sharpness-Aware Minimization (ASAM)~\cite{kwon_asam_2021} to bias training toward flatter minima, which are empirically associated with improved robustness to weight or input perturbations~\cite{stutz_relating_2021, keskar_large-batch_2017}. 
Hessian-Aware Pruning (HAP)~\cite{yu_hessian-aware_2022} uses curvature information to estimate parameter sensitivity.
Despite these advances, achieving strong corruption robustness at deployment-relevant high unstructured sparsity (e.g., 80--90\%) remains challenging.
First, because pruning constrains perturbations to the surviving subspace, a fixed SAM/ASAM radius can implicitly shrink the mean absolute perturbation over the full parameter vector (treating pruned coordinates as zeros) as sparsity increases.
Motivated by keeping the mean absolute perturbation (an $\ell_1$-based proxy) approximately invariant under sparsity, we derive a simple sparsity-dependent radius schedule that stabilizes sharpness-aware exploration during pruning.
Furthermore, second-order pruning methods such as HAP~\cite{yu_hessian-aware_2022} leverage Hessian information to estimate parameter sensitivity; however, at high sparsity the specific coupling between curvature and parameter scale can significantly affect which parameters are preserved.
As a simple complementary option, we evaluate Magnitude-Weighted Hessian (MWH), derived from a second-order removal-path analysis, yielding an importance proportional to $\mathrm{Diag}(F)_i\,|w_i|$, where $\mathrm{Diag}(F)$ is the diagonal empirical Fisher used as a curvature proxy in our implementation.
These limitations motivate stabilizing sharpness-aware exploration under sparsity, which we focus on via SA-SAM; we also study the impact of a simple curvature--magnitude importance variant (MWH) as an optional component.

\begin{figure}[!tb]
    \includegraphics[keepaspectratio, width=\linewidth]{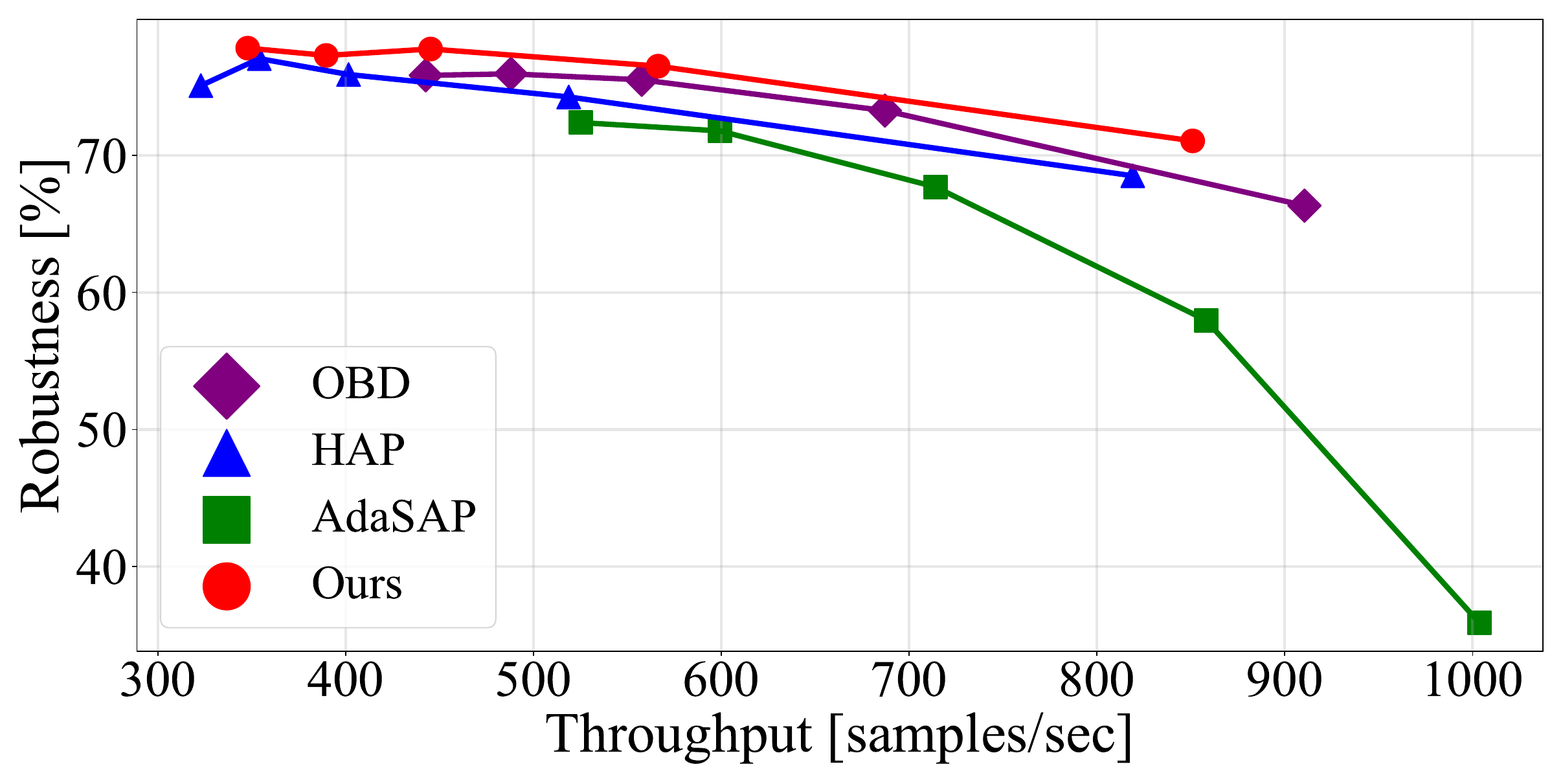}
    \caption{\textbf{Throughput vs.\ robustness on CIFAR-10-C.}
    The x-axis shows throughput (samples/sec) and the y-axis shows mean Top-1 accuracy (\%) on CIFAR-10-C averaged over all corruption types and severities. Throughput is averaged over 10 runs (full passes over the test set). Our method achieves higher corruption robustness than HAP and AdaSAP at comparable throughput (see legend). Note that at a fixed global sparsity, measured throughput can vary across methods due to differences in layer-wise nonzero distributions induced by different pruning criteria.
}
    \label{pareto}
\end{figure}

In this work, we propose Sparsity-Adaptive Sharpness-Aware Minimization~(SA-SAM) as an optimization method for pruning.
SA-SAM extends sharpness-aware training to sparse models by dynamically adjusting the perturbation radius according to the model's current sparsity, aiming to keep the effective exploration behavior stable throughout pruning.
Furthermore, we evaluate Magnitude-Weighted Hessian (MWH) as an optional importance score that combines curvature with weight magnitude in high-sparsity pruning.
We follow a standard gradual pruning pipeline that iteratively updates a pruning mask using an importance score and retrains the surviving weights; SA-SAM is a drop-in replacement for sharpness-aware optimization during pruning, and MWH is an optional drop-in replacement for the importance score.
As illustrated in Fig.~\ref{pareto}, our method improves the robustness--throughput trade-off on CIFAR-10-C at 80--90\% sparsity.
Across CIFAR-10-C, CIFAR-100-C, and ImageNet-100-C, we observe consistent robustness gains at these sparsity levels while maintaining competitive clean accuracy.
We further provide Fourier-domain and loss-landscape analyses (primarily on CIFAR-10) that help explain the observed robustness improvements.

Our contributions are summarized as follows:
\begin{itemize}
    \item \textbf{SA-SAM:} We propose a sparsity-adaptive sharpness-aware optimization method that calibrates the perturbation radius according to current sparsity, stabilizing sharpness-aware exploration during pruning.

    \item \textbf{Optional importance-score variant (MWH):} We evaluate a simple curvature--magnitude importance score, proportional to $\mathrm{Diag}(F)_i\,|w_i|$ (diagonal empirical Fisher as a curvature proxy), as an optional variant of second-order pruning criteria for high-sparsity robustness.

    \item \textbf{Robust high-sparsity pruning:} We demonstrate corruption-robustness gains at 80–90\% sparsity on CIFAR-10-C, CIFAR-100-C, and ImageNet-100-C, and provide a robustness--throughput analysis highlighting deployment-relevant trade-offs.
\end{itemize}
\textbf{Note.} The implementation code is publicly available at \url{https://github.com/ia-gu/Sparsity-Adaptive-Sharpness-Aware-Minimization}.

\section{Related Work}\label{sec:related}

\subsection{Pruning and robustness under common corruptions}
Pruning compresses neural networks by removing weights (unstructured) or groups of weights (structured) based on an importance score~\cite{cheng_survey_2024}.
Representative criteria include magnitude-based pruning~\cite{li_pruning_2017} and gradient-based saliency methods such as SNIP~\cite{lee_snip_2018}, as well as second-order criteria derived from local curvature.
While many pruning methods focus on preserving clean accuracy, robustness to distribution shifts and common corruptions (e.g., CIFAR-C/ImageNet-C~\cite{hendrycks_benchmarking_2018}) can degrade substantially at high sparsity.
This motivates incorporating robustness-aware objectives or sensitivity estimates into pruning pipelines.
Second-order pruning traces back to Optimal Brain Damage (OBD), which derives a saliency proportional to $H_{ii}w_i^2$ under a diagonal Hessian approximation~\cite{lecun_optimal_1989}.
Hessian-Aware Pruning (HAP)~\cite{yu_hessian-aware_2022} modernizes such ideas via efficient curvature estimation for pruning at scale.

\subsection{Sharpness-aware optimization and its variants}
Deep networks are typically optimized with first-order methods such as SGD or Adam, which minimize empirical risk.
However, minimizing training loss alone can converge to sharp minima that are sensitive to perturbations and may generalize poorly under distribution shifts.
Sharpness-Aware Minimization (SAM)~\cite{foret_sharpness-aware_2021} addresses this by optimizing parameters that minimize the worst-case loss within a neighborhood in \emph{weight space}, implemented by (i) applying an ascent perturbation toward the local worst direction and (ii) updating the original weights using gradients computed at the perturbed point.
Unlike adversarial training~\cite{zhao_adversarial_2022}, which targets worst-case perturbations in \emph{input space} and can induce a stronger clean-accuracy trade-off, SAM perturbs weights and is frequently observed to retain competitive clean accuracy while improving robustness in practice~\cite{zhang_duality_2024}.
Several variants improve SAM’s stability and efficiency.
ASAM~\cite{kwon_asam_2021} rescales perturbations based on parameter magnitudes to achieve approximate scale invariance, and SSAM~\cite{tan_stabilizing_2024} stabilizes updates by normalizing the sharpness-aware gradient to match the norm of the corresponding SGD gradient.
LookSAM~\cite{liu_towards_2022} reduce computational overhead via approximation or scheduling.
However, these methods are primarily designed for \emph{dense} models with a fixed effective dimension.
In gradual pruning, the feasible perturbations are constrained to the surviving subspace; with a fixed global radius, the average per-parameter perturbation magnitude can implicitly shrink as sparsity increases, weakening the sharpness-aware effect at deployment-relevant high sparsity.
Our SA-SAM directly targets this sparsity-induced shrinkage by adapting the global SAM/ASAM radius as a function of the current sparsity.

\section{Preliminaries}
\label{sec:preliminary}

\subsection{Sharpness-Aware Minimization}
\label{subsec:sam}

In image classification, given a training dataset $\mathcal{D}$ with input images $\mathbf{X}$ and labels $\mathbf{Y}$, a model learns by minimizing the loss $\ell$ with respect to parameters $\mathbf{W} \in \mathbb{R}^{m \times n}$ as follows:
\begin{equation}
    \min_{\mathbf{W}} \mathbb{E}_{(\mathbf{X}, \mathbf{Y}) \sim \mathcal{D}} \left[ \ell(f(\mathbf{X}; \mathbf{W}), \mathbf{Y}) \right]
\label{eq:loss}
\end{equation}
SAM~\cite{foret_sharpness-aware_2021} formulates learning as a bi-level optimization problem, minimizing the worst-case local loss in the weight space simultaneously:
\begin{equation}
    \min_{\mathbf W}
    \max_{\|\boldsymbol\epsilon\|_{2}\le \rho}
    \mathbb{E}_{(\mathbf X,\mathbf Y)\sim \mathcal{D}}
    \bigl[
        \ell\!\bigl(f(\mathbf X;\,\mathbf W+\boldsymbol\epsilon),\mathbf Y\bigr)
    \bigr],
\label{eq:sam}
\end{equation}
where $\rho>0$ is the radius of the perturbation ball~($\ell_2$-ball)~defined in the weight space, and it is a hyperparameter.
Using a first-order Taylor approximation of the inner objective, an approximate maximizer can be written as:
\begin{equation}
    \boldsymbol\epsilon^\star
    \;=\;
    \rho\,
    \frac{\nabla_{\mathbf W}\ell}{\|\nabla_{\mathbf W}\ell\|_2}
\label{eq:sam_ball}
\end{equation}
This enables an easy two-step implementation using standard optimizers, ``(i)~compute perturbation →~(ii)~update gradients.''
However, since SAM employs the same radius for all parameters, the sharpness is not invariant to weight re-scaling transformations $\mathbf W\!\mapsto\!a\mathbf W$, leading to scale-dependency issues.
ASAM~\cite{kwon_asam_2021} addresses this limitation by introducing a parameter-dependent re-scaling in the constraint. 
Specifically, the perturbation is constrained by:
\begin{equation}
    \|\mathbf T_{\mathbf W}^{-1}\,\boldsymbol\epsilon\|_{2}\;\le\;\rho,
    \qquad
    \mathbf T_{\mathbf W} \;=\; \mathrm{diag}\!\bigl(|\mathbf W|+\eta\bigr),
\label{eq:asam_tw}
\end{equation}
where $\eta>0$ is a small constant for numerical stability. The optimal perturbation then becomes:
\begin{equation}
    \boldsymbol\epsilon^\star
    \;=\;
    \rho\,
    \frac{\mathbf T_{\mathbf W}^{2}\,\nabla_{\mathbf W}\ell}
         {\bigl\|\mathbf T_{\mathbf W}\,\nabla_{\mathbf W}\ell\bigr\|_{2}}
 \label{eq:asam_ball}
\end{equation}
In this formulation, dimensions with larger parameter values admit proportionally larger perturbations, making the sharpness measure approximately invariant to weight re-scaling, thus alleviating the scale-dependency issue observed in SAM.

\subsection{Pruning}
\label{subsec:pruning_preliminary}

In pruning, since the mask $\mathbf{M} \in \{0,1\}^{m \times n}$ is applied to $\mathbf{W}$, Eq.~\eqref{eq:loss} can be written as:
\begin{equation}
    \min_{\mathbf{W}, \mathbf{M}} \mathbb{E}_{(\mathbf{X}, \mathbf{Y}) \sim \mathcal{D}} \left[ \ell(f(\mathbf{X}; \mathbf{W} \odot \mathbf{M}), \mathbf{Y}) \right]
\label{eq:loss_prune}
\end{equation}
Here, $\odot$ represents Hadamard product.  
In pruning, the importance score $I$ is designed to generate a mask satisfying the sparsity constraints.  
Denoting each element of $\mathbf{W}$ as $w_{i,j}$, the mask matrix is defined by calculating the importance $I(w_{i,j})$ for each parameter:
\begin{equation}
M_{i,j} =
\begin{cases}
1, & I(w_{i,j}) \geq \tau \\
0, & I(w_{i,j}) < \tau
\end{cases}
\label{eq:mask}
\end{equation}
We choose $\tau$ such that exactly the top $(1-s)mn$ parameters (by $I$) are kept.

\section{Method}
\subsection{Sparsity-Adaptive Sharpness-Aware Minimization}
Let $W \in \mathbb{R}^d$ denote the vectorized \emph{prunable} weight vector
(convolution and linear weights) and let $m \in \{0, 1\}^d$ be the corresponding
elementwise pruning mask; we write $M = \mathrm{diag}(m)$ so that masking is
$MW = W \odot m$.
BatchNorm parameters and biases are kept dense in all experiments and are omitted
from $W$ in this derivation for simplicity.
Let $S = \{i : m_i = 1\}$ be the support with $d_s = |S| = (1 - s)d$,
where $s$ denotes the sparsity over these prunable weights, and let $\Pi_S = M$
be the projector onto the coordinate subspace $\mathrm{range}(\Pi_S) \subset \mathbb{R}^d$.
The inner maximization must be solved on that subspace.
Throughout this subsection, we abuse notation and write $\ell(W)$ as shorthand for $\ell(f(X; W \odot m), Y)$ when the pruning mask $m$ is fixed.
To avoid inverting a singular matrix outside $S$, we use the Moore–Penrose pseudoinverse:
\begin{equation}
\max_{\epsilon_S\in\mathrm{range}(\Pi_S)}\;
\left\langle \nabla_W \ell(W),\,\epsilon_S \right\rangle
\quad \text{s.t.}\quad
\left\| (\,\Pi_S T_W \Pi_S\,)^{+}\epsilon_S\right\|_2 \le \rho .
\label{eq:inner}
\end{equation}
Here $T_W$ denotes the diagonal ASAM scaling matrix; for SAM, set $T_W=I$.
Solving the Lagrangian of the linearized inner problem yields the closed-form perturbation in $\mathrm{range}(\Pi_S)$:
\begin{equation}
\epsilon^\star
= \rho\;\frac{\Pi_S T_W^2 \nabla_W\ell}{\|\Pi_S T_W \nabla_W\ell\|_2},
\label{eq:inner-solution}
\end{equation}
which coincides with the heuristic formula obtained by simply zeroing pruned coordinates, but Eq.~\eqref{eq:inner} and Eq.~\eqref{eq:inner-solution} make the subspace and invertibility explicit.

Pruning reduces the dimension $d\!\to\! d_s$ and can shrink the per-parameter step implied by Eq.~\eqref{eq:inner-solution}. 
We quantify the global step size by the mean absolute step $\Phi(\epsilon)=\tfrac{1}{d}\|\epsilon\|_1$, which measures the average absolute perturbation per parameter in the \emph{original} dense parameterization.
This choice makes the effective sharpness-aware exploration comparable across sparsity levels when pruned coordinates are treated as zeros.
Let $t=\mathrm{diag}(T_W)\in\mathbb{R}^d_{+}$ and define the averages $\,\bar t=\tfrac{1}{d}\sum_i t_i\,$ and $\,\bar t_S=\tfrac{1}{d_s}\sum_{i\in S} t_i\,$.
Write $u=\frac{\Pi_S T_W\nabla_W\ell}{\|\Pi_S T_W\nabla_W\ell\|_2}\in\mathbb{S}^{d_s-1}$.
Under the standard exchangeability/isotropy approximation for $u$,
$\mathbb{E}[|u_1|]=c(d_s)\approx\sqrt{2/(\pi d_s)}$,
and we obtain
\begin{equation}
\mathbb{E}\!\left[\Phi(\epsilon^\star)\mid s\right]
= \rho\,(1-s)\,\bar t_S\,c(d_s).
\label{eq:mean-step}
\end{equation}
Relative to the dense case ($s=0$), the ratio is
\begin{equation}
\frac{\mathbb{E}[\Phi(\epsilon^\star)\mid s]}{\mathbb{E}[\Phi(\epsilon^\star)\mid 0]}
= (1-s)\,\frac{\bar t_S}{\bar t}\,\frac{c(d_s)}{c(d)}
\;\approx\; \sqrt{1-s}\cdot\frac{\bar t_S}{\bar t}.
\label{eq:ratio}
\end{equation}
To maintain invariant step size, we set
\begin{align}
\rho^\star(s)&=\rho\cdot \alpha(s), \nonumber\\
\quad
\alpha(s):&=\frac{c(d)}{(1-s)\,c(d_s)}\cdot\frac{\bar t}{\bar t_S}
\;\approx\;\frac{1}{\sqrt{1-s}}.
\label{eq:alpha}
\end{align}
In practice we use the simple approximation, which depends only on $s$ (available from the mask) and introduces no extra schedules. 
For SAM, set $T_W=I$ in Eqs.~\eqref{eq:inner}--\eqref{eq:alpha}.

\textbf{Remarks.}
The exchangeability/isotropy assumption for $u$ provides a tractable approximation of typical coordinate magnitudes; in practice, we found that using the approximation $\alpha(s) \approx 1/\sqrt{1-s}$ works well.
Calibrating $\rho$ with Eq.~\eqref{eq:alpha} (or its approximation) compensates for the reduced effective dimension and mitigates the shrinkage of sharpness-aware perturbations at high sparsity.

\subsection{Optional importance-score: Magnitude-Weighted Hessian}
In our experiments, we approximate the Hessian diagonal $H_{ii}$ with a non-negative curvature proxy, namely the diagonal empirical Fisher $\mathrm{Diag}(F)_i$, and we use the term ``curvature'' to refer to this proxy.
Magnitude-based methods rank parameters by $|w_i|$, while curvature-based methods use second-order information.
Following a second-order removal-path analysis, we consider removing a weight along $w_i(t)=(1-t)w_i,\;t\in[0,1]$, and use the quadratic expansion
\begin{equation}
\Delta\ell(t) \approx -t\,w_i\,\nabla_i\ell \;+\; \tfrac{1}{2}\,t^2\,H_{ii}\,w_i^2.
\label{eq:path-expansion}
\end{equation}
Assuming the first-order term is small (as in OBD), the second-order approximation predicts the loss increase from full removal ($t=1$) as $\Delta\ell(1)\approx \tfrac{1}{2}H_{ii}w_i^2$.
Normalizing by the removal-path length $|w_i|$ yields the per-unit-removal penalty
\begin{equation}
\frac{1}{|w_i|}\cdot \frac{1}{2} H_{ii}|w_i|^2
=\frac{1}{2}\,H_{ii}\,|w_i|
\;\propto\; H_{ii}\,|w_i|.
\label{eq:mwh}
\end{equation}

\textbf{Curvature proxy (diagonal Fisher).}
In our implementation, we use the diagonal empirical Fisher as a practical, non-negative proxy for curvature.
Concretely, when computing the score we replace $H_{ii}$ in Eq.~\eqref{eq:mwh} with $\mathrm{Diag}(F)_i$.
Let $g(\mathbf{x},\mathbf{y})=\nabla_W \ell(f(\mathbf{x};W),\mathbf{y})$ denote the gradient on a sample.
Let $g_b=\nabla_W \ell(f(x_b;W),y_b)$ denote the gradient of the per-example loss for a sample $b$.
We estimate the diagonal empirical Fisher using squared gradients computed at mask-update time.
(A) If per-example gradients $\{g_b\}_{b\in B}$ are available, we estimate $\mathrm{Diag}(F)$ by averaging the elementwise squared per-example gradients over a single mini-batch $B$.
(B) Otherwise, we use the squared mini-batch gradient as a proxy, i.e., $\mathrm{Diag}(F) := g_B \odot g_B$ where $g_B = \nabla_W \frac{1}{|B|}\sum_{b\in B}\ell_b$.
In all experiments reported in this paper, we use option (B), i.e., the squared mini-batch gradient proxy, for efficiency.
To keep the curvature proxy consistent across pruning iterations, we use a single fixed mini-batch $B$ (sampled once per run/seed) and reuse it for all mask updates.

Discarding constant factors, our MWH importance score is
\begin{equation}
I_i \;=\; \mathrm{Diag}(F)_i \cdot |w_i|.
\label{eq:mwh-score}
\end{equation}
Given a target sparsity $s$, we keep the top $(1-s)$ fraction ranked by Eq.~\eqref{eq:mwh-score} and zero the rest:
\begin{equation}
m_i = \mathbf{1}\big[I_i \text{ is among the top } (1-s)d\big],\qquad
W \leftarrow W \odot m .
\label{eq:mask-update}
\end{equation}
Computationally, SA-SAM has the same per-step overhead as ASAM, since it only rescales the global radius as a function of sparsity, and MWH reuses squared-gradient statistics computed at mask-update time.

\begin{table*}[t]
\centering
\caption{\textbf{CIFAR-10 / CIFAR-100.} Top-1 (\%) on the clean test sets and on the corruption sets (mean over 19 corruption types $\times$ 5 severities). Values are mean $\pm$ std over 3 seeds. Dense denotes the baseline with sparsity $s{=}0.0$. The rightmost column reports inference throughput (samples/sec).
\textbf{Ours} uses SA-SAM during pruning retraining and MWH for mask updates.}
\label{tab:cifar}
{
\begin{tabular}{lcccc|c}
\toprule
Method@$s$ & CIFAR-10{} & CIFAR-10{}-C & CIFAR-100{} & CIFAR-100{}-C & Throughput \\
\midrule
Dense@0.0  & 96.12 $\pm$ 0.03 & 78.19 $\pm$ 0.46 & 79.93 $\pm$ 0.05 & 53.37 $\pm$ 0.44 & 200 $\pm$ 0 \\
\midrule
OBD@0.8     & 93.86 $\pm$ 0.10 & 75.82 $\pm$ 0.54 & 76.50 $\pm$ 0.30 & 48.61 $\pm$ 0.41 & 446 $\pm$ 11 \\
HAP@0.8     & 95.29 $\pm$ 0.12 & 75.08 $\pm$ 0.80 & 78.46 $\pm$ 0.21 & 48.97 $\pm$ 0.66 & 316 $\pm$ 6 \\
AdaSAP@0.8  & 93.59 $\pm$ 0.16 & 72.38 $\pm$ 0.96 & 72.72 $\pm$ 0.13 & 45.64 $\pm$ 0.71 & 689 $\pm$ 12 \\
Ours@0.8& \textbf{96.00 $\pm$ 0.07} & \textbf{77.81 $\pm$ 0.53} & \textbf{79.49 $\pm$ 0.05} & \textbf{52.63 $\pm$ 0.53} & 535 $\pm$ 7 \\
\midrule
OBD@0.9     & 93.64 $\pm$ 0.03 & 75.50 $\pm$ 0.53 & 75.19 $\pm$ 0.37 & 46.32 $\pm$ 0.57 & 1310 $\pm$ 90 \\
HAP@0.9     & 95.76 $\pm$ 0.08 & 75.88 $\pm$ 0.81 & 77.01 $\pm$ 0.54 & 47.22 $\pm$ 0.76 & 1443 $\pm$ 118 \\
AdaSAP@0.9  & 89.62 $\pm$ 0.13 & 67.69 $\pm$ 0.81 & 61.41 $\pm$ 0.09 & 37.70 $\pm$ 0.46 & 1680 $\pm$ 167 \\
Ours@0.9& \textbf{96.12 $\pm$ 0.10} & \textbf{77.74 $\pm$ 0.53} & \textbf{78.80 $\pm$ 0.14} & \textbf{51.16 $\pm$ 0.56} & 1423 $\pm$ 116 \\
\bottomrule
\end{tabular}}
\vspace{-2mm}
\end{table*}

\begin{table}[t]
\centering
\caption{\textbf{ImageNet-100.} Top-1 (\%) on the clean test set and the corruption set.}
\label{tab:imagenet100}
\resizebox{\columnwidth}{!}{
\begin{tabular}{lcc|c}
\toprule
Method@$s$ & ImageNet-100{} & ImageNet-100{}-C & Throughput \\
\midrule
Dense@0.0    & 82.91 $\pm$ 0.30 & 44.58 $\pm$ 0.69 & 173 $\pm$ 3 \\
\midrule
OBD@0.8      & 81.81 $\pm$ 0.42 & 40.86 $\pm$ 1.18 & 365 $\pm$ 3 \\
HAP@0.8      & 81.27 $\pm$ 0.53 & 42.13 $\pm$ 0.53 & 381 $\pm$ 14 \\
AdaSAP@0.8   & 77.51$
\pm$0.57 & 35.74 $\pm$ 0.82 & 426 $\pm$ 17 \\
Ours@0.8 & \textbf{83.14 $\pm$ 0.06} & \textbf{44.45 $\pm$ 0.50} & 387 $\pm$ 15 \\
\midrule
OBD@0.9      & 80.93 $\pm$ 0.56 & 39.59 $\pm$ 1.47 & 420 $\pm$ 17 \\
HAP@0.9      & 80.63 $\pm$ 0.36 & 39.36 $\pm$ 1.21 & 442 $\pm$ 19 \\
AdaSAP@0.9   & 72.30 $\pm$ 0.47 & 31.27 $\pm$ 0.90 & 512 $\pm$ 25 \\
Ours@0.9 & \textbf{82.80 $\pm$ 0.16} & \textbf{43.17 $\pm$ 0.47} & 449 $\pm$ 19 \\
\bottomrule
\end{tabular}}
\vspace{-2mm}
\end{table}

\section{Experiments}\label{sec:experiment}
\subsection{Settings}\label{subsec:settings}
\textbf{Datasets.} We evaluate on CIFAR-10, CIFAR-100~\cite{krizhevsky_learning_2009} and ImageNet-100~\cite{shekhar_imagenet100_2021}; robustness is assessed on CIFAR-10-C, CIFAR-100-C and ImageNet-100-C~\cite{hendrycks_benchmarking_2018}.
ImageNet-100 is a 100-class subset of ImageNet~\cite{deng_imagenet_2009} used to reduce computational cost; ImageNet-100-C is constructed by restricting ImageNet-C to the same 100 classes using the synset list of ImageNet-100; concretely, we filter the ImageNet-C validation images by the selected synset IDs and then evaluate using the standard ImageNet-C corruption protocol.
Unless otherwise noted, we report Top-1 accuracy on the clean test set and mean Top-1 accuracy over all corruptions and severities.
For CIFAR-10-C and CIFAR-100-C we average over 19 corruption types and 5 severity levels, while for ImageNet-100-C we follow the ImageNet-C protocol (15 corruption types, 5 severity levels).

\textbf{Architectures.} ResNet-18 for CIFAR-10 and CIFAR-100, ResNet-50 for ImageNet-100. 
\textbf{Methods compared.}
We study two orthogonal design choices in gradual pruning:
(i) the sharpness-aware optimizer used during the retraining steps (ASAM vs.\ SA-SAM),
and (ii) the importance score used for mask updates (OBD/HAP/AdaSAP vs.\ MWH).
Unless otherwise stated, \textbf{Ours} denotes \textbf{SA-SAM + MWH}, i.e., we apply SA-SAM during pruning retraining and use MWH for mask updates.
Unless otherwise stated, \textbf{Ours} denotes \textbf{SA-SAM + MWH}.
To isolate the optimizer effect, we additionally report ASAM $\rightarrow$ SA-SAM swaps while keeping the pruning criterion fixed (Table~\ref{tab:ablation_in}).
For MWH, we use the diagonal empirical Fisher (approximated by a squared mini-batch gradient; see Sec.~IV-B) as a practical curvature proxy and rank parameters by $\mathrm{Diag}(F)_i\,|w_i|$ (Eq.~\eqref{eq:mwh-score}).
To isolate the effect of sparsity-adaptive radius scaling, we keep the base optimizer and training schedule fixed across methods; SA-SAM introduces no additional tunable hyperparameters beyond the initial radius $\rho_0$ used in ASAM.

We emphasize that our experimental focus is corruption robustness under deployment-relevant high unstructured sparsity.
We therefore compare against pruning pipelines that explicitly incorporate curvature-aware mechanisms in this context (e.g., AdaSAP and HAP), along with a classical diagonal-Hessian baseline (OBD).
A broader comparison against generic saliency criteria (e.g., Fisher/Taylor-style variants) is beyond the scope of this paper and left as future work.

\textbf{Optimization.} All methods use learning rate 0.1, momentum 0.9, weight decay 0.0005, batch size 128, and a base sharpness radius of $\rho_0 = 0.5$.
For SA-SAM, we set $\rho(s) = \rho_0 \alpha(s)$ with $\alpha(s) \approx 1/\sqrt{1-s}$ following Eq.~\eqref{eq:alpha}.
We use a step learning-rate schedule that multiplies the learning rate by 0.2 at epochs 120 and 160.
Dense, HAP and AdaSAP use ASAM; our method uses SA-SAM with the same base optimizer and hyperparameters.
For ASAM, we use the standard stabilizer $\eta=0.01$ in $\mathbf{T}_{\mathbf{W}}=\mathrm{diag}(|\mathbf{W}|+\eta)$.

\textbf{Pruning.} For all pruning methods, we perform gradual pruning for 220 epochs. At each epoch, we update the mask to match a target sparsity that increases according to a fixed schedule until reaching the final sparsity $s\in\{0.8,0.9\}$.
Specifically, at epoch $t \in \{1, \ldots, T\}$ we set $s_t = s_{\mathrm{final}} \cdot (t/T)^3$.
At the beginning of each epoch, we compute the importance scores on the current model, update the mask to keep the top $(1 - s_t)$ fraction, and then train for one epoch while keeping the updated mask fixed.
We prune convolution and linear weights with unstructured masks.
Throughout the paper, the reported global sparsity $s$ is computed over the prunable convolution and linear weights only; BatchNorm parameters and biases are excluded.
For a fair comparison, we run all methods under the same gradual-pruning schedule, training budget (epochs), data pipeline, and base optimizer settings; only the sharpness-aware optimizer (ASAM vs.\ SA-SAM) and/or the importance score differ as specified.

\textbf{Throughput measurement.}
We measure inference throughput (samples/sec) on an NVIDIA RTX 3090.
All throughput numbers are measured under the same COO sparse-execution backend for consistency across sparsity levels; Dense@0.0 is included as a reference under this backend.
We report mean $\pm$ std over 10 runs (after warm-up) with batch size 128.
Exact software versions and implementation details will be provided in the released code.
Throughput numbers are backend- and batch-size-dependent; we report them to provide a controlled robustness--throughput comparison under a single COO sparse-execution backend, rather than as absolute edge-device latency.

Unless otherwise stated, Tables~I--II use ASAM for Dense/HAP/AdaSAP and SA-SAM for Ours; Section~V-D additionally reports ASAM vs.\ SA-SAM ablations for each pruning method.

\subsection{Experimental Results}\label{subsec:results}
\begin{figure}[!tb]
    \centering
    \includegraphics[keepaspectratio, width=\linewidth]{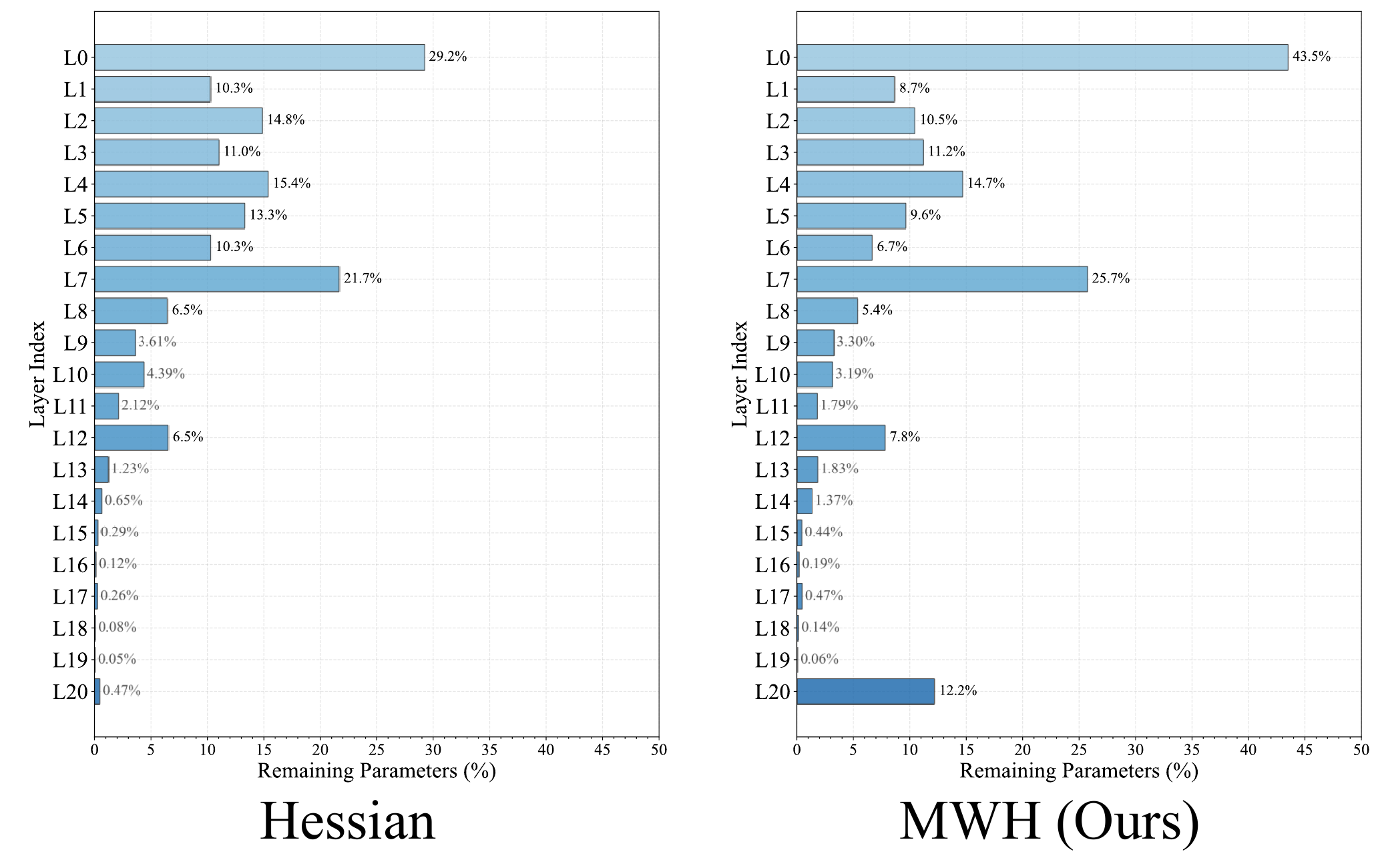}
    \caption{\textbf{Layer-wise remaining parameter ratios at a fixed global sparsity ( ResNet-18 on CIFAR-10 at $s=0.9$).} }
    \label{fig:layerwise}
\end{figure}

Tables~\ref{tab:cifar} and \ref{tab:imagenet100} summarize clean and corruption accuracy at $s{=}0.8,0.9$.
Overall, \textbf{Ours (SA-SAM + MWH)} improves corruption robustness over HAP and AdaSAP while preserving clean Top-1.
The gains are more pronounced at $s=0.9$: on CIFAR-10-C, we improve mean corruption accuracy from 75.88 (HAP) / 67.69 (AdaSAP) to 77.74 while matching clean accuracy (96.12), and achieve comparable throughput to HAP (1423 vs.\ 1443 samples/sec).
On CIFAR-100-C at $s=0.9$, we improve corruption accuracy from 47.22 (HAP) / 37.70 (AdaSAP) to 51.16.
On ImageNet-100-C at $s=0.9$, we improve corruption accuracy from 39.36 (HAP) / 31.27 (AdaSAP) to 43.17 with clean Top-1 of 82.80.
We also report measured throughput in Tables~\ref{tab:cifar}--\ref{tab:imagenet100}; because different pruning criteria induce different layer-wise nonzero patterns, throughput can differ even at the same global sparsity (see Fig.~\ref{fig:layerwise}).
To disentangle the contributions of the optimizer and the pruning criterion, Table~\ref{tab:ablation_in} further reports optimizer-only swaps (ASAM $\rightarrow$ SA-SAM) on ImageNet-100 and shows consistent gains in corruption robustness, while indicating that MWH can also help under a fixed optimizer.

\subsection{Analysis}\label{subsec:analysis}
\textbf{Fourier heat map.}
Fourier heat map analysis examines model robustness from a frequency domain perspective.
We perturb test images using single-frequency Fourier-basis noise and measure the classification error.
Fourier heat map is evaluated both qualitatively through visualization and quantitatively using the performance degradation rate $D_{i, j}$:
\begin{equation}
D_{i,j} = \frac{E_{\text{pruned}(i, j)} - E_{\text{dense}(i, j)}}{E_{\text{dense}(i, j)}} \times 100 \quad (\%)
\label{eq:errr_degradation}
\end{equation}
Finally, the robustness of each model is obtained by calculating $D_{i, j}$ for all coordinates and then averaging them.
Furthermore, to examine the tendency for each frequency band, we define the range satisfying $|i|+|j| \le 8$ as ``Low'' frequency and the rest as ``High'' frequency.

\begin{figure}[!tb]
    \begin{center}
    \includegraphics[keepaspectratio, width=\linewidth]{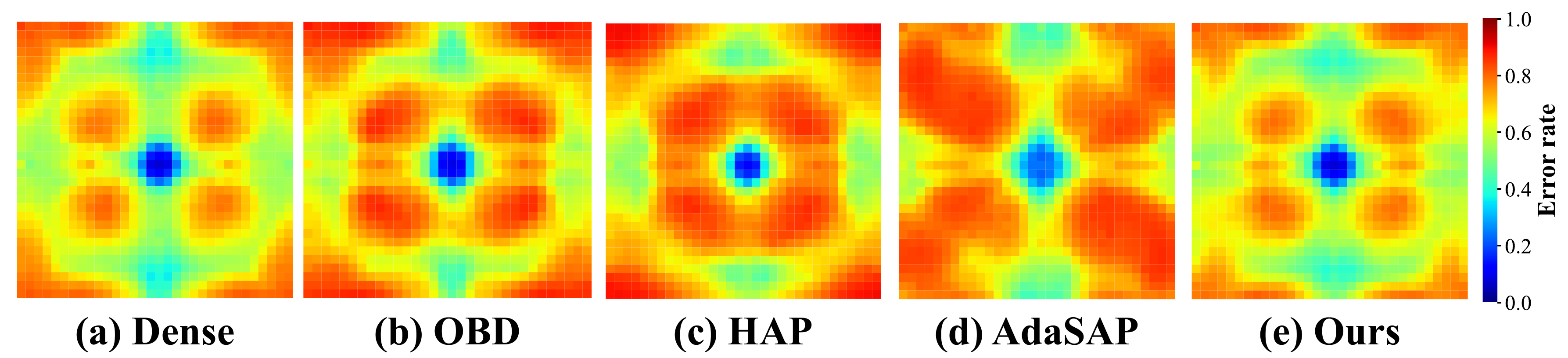}
    \caption{\textbf{Fourier heat maps on CIFAR-10} for (a) Dense, (b) OBD, (c) HAP, (d) AdaSAP, and (e) Ours. Methods (b)-(e) used sparsity $s=0.9$. Our method shows smaller degradation in the low-frequency region compared to other sparse baselines.
}
    \label{fhm}
    \end{center}
\end{figure}

Fig.~\ref{fhm} visualizes the Fourier heat maps, and Table~\ref{tab:fhm} reports the corresponding degradation rates.
The central area indicates low-frequency performance, while the outer area shows high-frequency performance. Blue indicates higher performance, red indicates lower performance.
The table shows degradation rates for Low, High, and Average frequencies, where smaller values indicate better maintenance of Dense performance.
From the figure and the table, our method improved the model robustness over the baselines.
While our Fourier-domain analysis is conducted on CIFAR-10 (32$\times$32), its trend---notably the smaller average degradation in Table~\ref{tab:fhm}---was consistent with the end-to-end robustness gains observed on ImageNet-100-C in Table~\ref{tab:imagenet100}.
Many common corruptions in ImageNet-100-C contain significant high-frequency components, and the improved frequency-wise stability we observe on CIFAR may be related to the end-to-end robustness gains at larger resolution; verifying this with a systematic Fourier analysis on ImageNet-100 is future work.

\begin{table}[tb!]
\centering
\caption{Frequency-wise performance degradation for the sparse models ($s=0.9$). Each value is the degradation rate defined in Eq.~\eqref{eq:errr_degradation}, lower is better.}
\begin{tabular}{l c c c c}
\toprule
Method  & Low↓ & High↓ & Average↓ \\
\midrule
OBD & 9.45 & 3.80 & 5.02 \\
HAP & 10.45 & 3.52 & 5.01 \\
AdaSAP  & 12.46 & 15.29 &14.68 \\
Ours & \bf{1.94} & \bf{0.13} & \bf{0.52} \\
\bottomrule
\end{tabular}
\label{tab:fhm}
\end{table}

\textbf{Loss landscape and Eigenvalue Spectrum Density.}
Loss landscape~\cite{li_visualizing_2018} visualizes the geometry around a model’s solution by applying random perturbations to the trained parameters and plotting the resulting loss (or error rate).
We perturb the trained weights along two random directions and plot the resulting error-rate surface; we use the same plotting protocol for all methods.
We evaluate the geometry quantitatively via the Hessian Eigenvalue Spectrum Density (ESD)~\cite{yao_pyhessian_2020}.
ESD characterizes the distribution of Hessian eigenvalues around a trained solution.
We do not form the full Hessian explicitly; instead, we estimate the spectrum using Hessian--vector products and a Lanczos-based procedure following standard practice.
We approximate the top 100 eigenvalues $\{\lambda_i\}_{i=1}^{100}$ via Lanczos iterations and report the $\ell_2$ energy $\left(\sum_{i=1}^{100}\lambda_i^2\right)^{1/2}$ as a proxy for curvature magnitude.

\begin{figure}[!tb]
    \begin{center}
    \includegraphics[keepaspectratio, width=\linewidth]{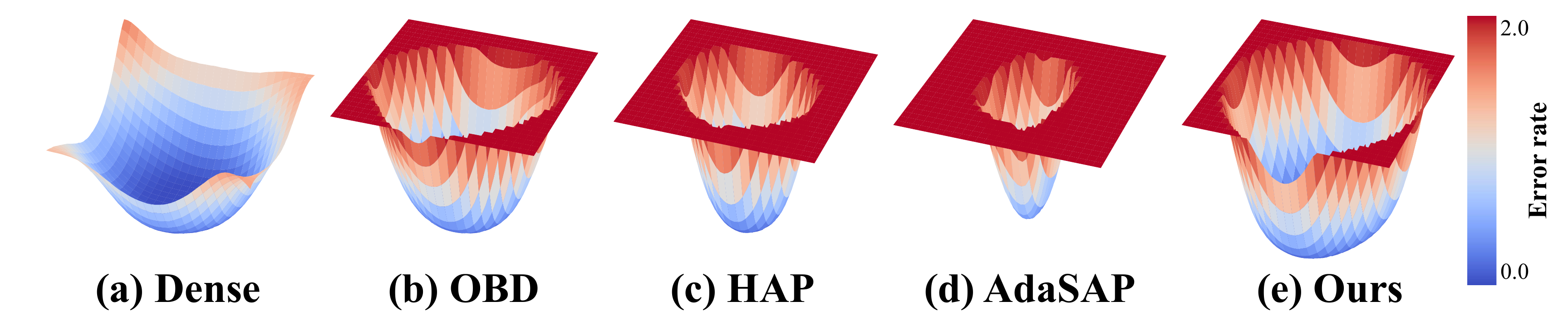}
    \caption{\textbf{Loss landscapes on CIFAR-10} for (a) Dense, (b) OBD, (c) HAP, (d) AdaSAP, and (e) Ours. Methods (b)-(e) use sparsity $s=0.9$. Each surface is obtained by applying perturbations along two random directions in weight space and plotting the resulting error rate.}
    \label{ll}
    \end{center}
\end{figure}

\begin{figure}[!tb]
    \begin{center}
    \includegraphics[keepaspectratio, width=\linewidth]{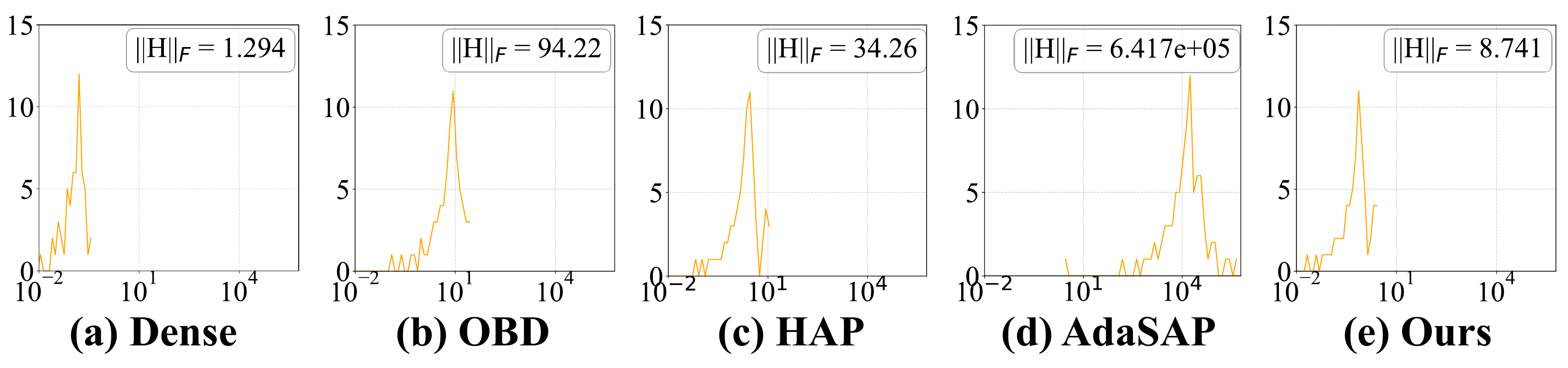}
    \caption{\textbf{ESDs ($\mathbf{\lambda}$=100) on CIFAR-10} for (a) Dense, (b) OBD, (c) HAP, (d) AdaSAP, and (e) Ours. Methods (b)-(e) use sparsity $s=0.9$. The x-axis shows eigenvalues and the y-axis shows density. The $\ell_2$ energy of the top-100 eigenvalues, $\bigl(\sum_{i=1}^{100}\lambda_i^2\bigr)^{1/2}$, is reported in the top-right corner of each plot.}
    \label{esd}
    \end{center}
\end{figure}

Fig.~\ref{ll} and Fig.~\ref{esd} show Loss Landscapes and ESDs, respectively.
From~Fig.~\ref{ll}, our method yielded a flatter loss landscape than existing methods, which means that our model performs well on both the clean and corruption sets, similar to the results in Tables~\ref{tab:cifar} and~\ref{tab:imagenet100}.
From~Fig.~\ref{esd}, we observed the same tendency.
The reported value ($\|\mathbf{H}\|_F$) indicates that our model yields the smallest curvature energy among the sparse models.

\begin{table*}[t]
\centering
\caption{\textbf{Comparison of ASAM with SA-SAM for ImageNet-100.}
Top-1 (\%) on the clean test set and ImageNet-100-C (averaged over 15 corruption types $\times$ 5 severities).
Throughput is omitted because all settings use the same target sparsity and unstructured masks, and throughput differences across optimizers are negligible in this ablation.}
\label{tab:ablation_in}
\resizebox{\textwidth}{!}{
\begin{tabular}{lccccc}
\toprule
& & \multicolumn{2}{c}{ImageNet-100} & \multicolumn{2}{c}{ImageNet-100-C} \\
\cmidrule(lr){3-4} \cmidrule(lr){5-6}
Sparsity & Method & ASAM & SA-SAM & ASAM & SA-SAM \\
\midrule
0.00 & Dense & 82.91  $\pm$  0.30 & 82.91  $\pm$  0.30 & 44.58  $\pm$  0.69 & 44.58  $\pm$  0.69 \\
\midrule
\multirow{4}{*}{0.80} & OBD & $81.81\pm0.42$ & $81.37\pm 0.31$ (\color{blue}{$\downarrow0.44$)} & $40.86\pm 1.18$ & $40.34\pm 1.13$ (\color{blue}{$\downarrow0.52$})\\
& HAP & $81.27\pm0.53$ & $82.08\pm 0.46$ (\color{red}{$\uparrow0.81$)} & $42.13\pm 0.53$ & $42.32\pm 0.62$ (\color{red}{$\uparrow0.19$})\\
& AdaSAP & $77.51\pm 0.57$ & $77.82\pm 0.17$ (\color{red}{$\uparrow0.31$)} & $35.74\pm 0.82$ & $36.29\pm 0.73$ (\color{red}{$\uparrow0.55$}) \\
& MWH (Ours) & $82.85\pm 0.42$ & \textbf{83.14  $\pm$  0.06} (\color{red}{$\uparrow0.29$)}& $44.04\pm 0.57$ &  \textbf{44.45  $\pm$  0.50} (\color{red}{$\uparrow0.41$})\\
\midrule
\multirow{4}{*}{0.90} & OBD & $80.93\pm0.56$ & $81.33\pm 0.33$ (\color{red}{$\uparrow0.40$)} & $39.59\pm 1.47$ & $39.84\pm 1.03$ (\color{red}{$\uparrow0.25$})\\
& HAP & $80.63\pm 0.36$ & $80.71\pm 0.13$ (\color{red}{$\uparrow0.08$)} & $39.36\pm 1.21$ & $40.16\pm 0.64$ (\color{red}{$\uparrow0.80$}) \\
& AdaSAP & $72.30\pm 0.47$ & $72.58\pm 0.50$ (\color{red}{$\uparrow0.28$)}& $31.27\pm 0.90$ & $31.83\pm 0.74$ (\color{red}{$\uparrow0.56$}) \\
& MWH (Ours) & $82.27\pm 0.28$ & \textbf{82.80}  $\pm$  \textbf{0.16}  (\color{red}{$\uparrow0.53$)}& $42.88\pm 0.44$ & \textbf{43.17}  $\pm$  \textbf{0.47} (\color{red}{$\uparrow0.29$}) \\
\bottomrule
\end{tabular}}
\vspace{-2mm}
\end{table*}

\subsection{Ablations}
To demonstrate the effectiveness of the proposed optimizer SA-SAM, we compare ASAM with SA-SAM for all pruning methods.
For each method, we keep the pruning criterion, pruning schedule, and training budget fixed, and only replace the sharpness-aware optimizer (ASAM $\rightarrow$ SA-SAM) by scaling the global radius using Eq.~\eqref{eq:alpha}.
Experimental settings are the same as the main settings.
We omit throughput because all settings use the same target sparsity and unstructured masks; throughput differences across optimizers are negligible and not the focus of this ablation.

Table~\ref{tab:ablation_in} shows the result of the ablation experiments. 
As shown in the table, SA-SAM consistently improves (or matches within variance) clean and corruption accuracy across pruning methods, with more pronounced gains in corruption robustness at higher sparsity.
Under ASAM, MWH yields the strongest overall performance among the compared pruning criteria.
These results support the effectiveness of SA-SAM and the complementary benefit of MWH under high sparsity.

\begin{figure}[!tb]
    \begin{center}
    \includegraphics[keepaspectratio, width=\linewidth]{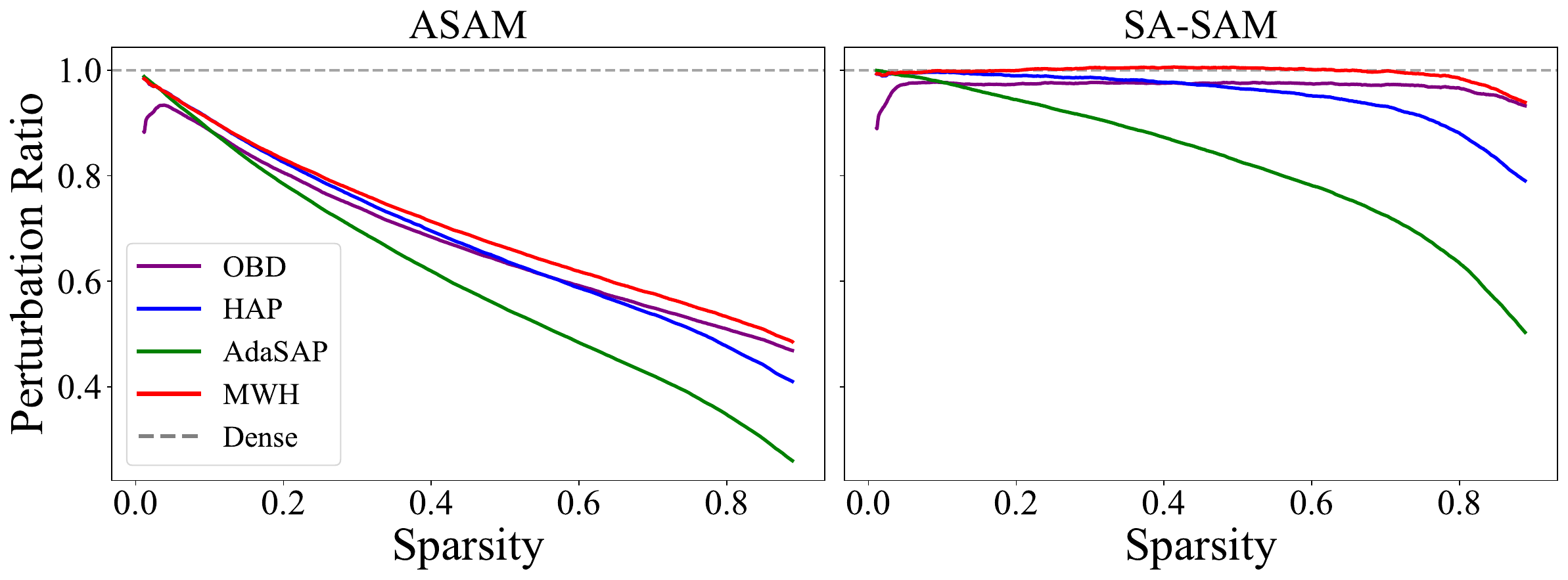}
    \caption{\textbf{Perturbation magnitude ratio} relative to the dense model across different pruning methods. The gray dashed line represents the dense model.}
    \label{fig:perturbation_ratio}
    \end{center}
\end{figure}

Finally, to directly validate the motivation of SA-SAM, we measure the mean absolute perturbation $\bar{\Phi}(\boldsymbol{\epsilon}^*) = \|\boldsymbol{\epsilon}^*\|_1/d$ and report its ratio to the dense baseline as pruning progresses. 
Fig.~\ref{fig:perturbation_ratio}, (left) shows that under ASAM with a fixed radius, the perturbation magnitude shrinks substantially as sparsity increases, consistent with the sparsity-induced shrinkage predicted by Eq.~\eqref{eq:ratio}.
In contrast, SA-SAM rescales the radius as Eq.~\eqref{eq:alpha}, which compensates for the reduced effective dimension and maintains perturbation magnitudes closer to the dense baseline across all pruning criteria (Fig.~\ref{fig:perturbation_ratio}, right). 
This confirms that SA-SAM preserves the intended sharpness-aware exploration behavior throughout gradual pruning.

\section{Conclusion}\label{sec:conclusion}
In this work, we proposed SA-SAM, which adapts the ASAM optimization method, originally designed for dense models, to sparse models created by pruning.
As a complementary option, we also evaluated MWH, a simple curvature--magnitude importance variant, to study its benefit under high sparsity.
Through experiments using multiple datasets and models with different parameter counts, we demonstrated the effectiveness of our proposed methods. 
Furthermore, through detailed analysis on CIFAR-10, we clarified the factors behind the performance improvement over existing methods.
Future work includes extending SA-SAM to structured sparsity and validating sparse execution across more optimized kernels and hardware backends.

\section*{Acknowledgments}
We used a GPT-based language model (ChatGPT, OpenAI) for minor English language polishing.
We verified all technical content, experimental settings, and results.

\bibliographystyle{IEEEtran}
\bibliography{revised}

\end{document}